\documentclass[letterpaper]{article} 
\usepackage{aaai2026}  
\usepackage{times}  
\usepackage{helvet}  
\usepackage{courier}  
\usepackage[hyphens]{url}  
\usepackage{graphicx} 
\usepackage{natbib}  
\usepackage{caption} 
\usepackage{algorithm}
\usepackage{algorithmic}
\usepackage{booktabs} 
\usepackage{float}
\usepackage{subfigure}
\usepackage{cuted}
\usepackage{capt-of}
\usepackage{amssymb} 

\usepackage[table]{xcolor}
\definecolor{lightpink}{rgb}{1,0.88,0.88} 
\definecolor{lightyellow}{rgb}{1,1,0.8}   
\makeatletter
\renewcommand{\@fnsymbol}[1]{\ifcase#1\or $\dagger$\or $\ddagger$\or
   $\mathsection$\or $\mathparagraph$\or $\|$\or **\or $\dagger\dagger$
   \or $\ddagger\ddagger$\fi}
\makeatother
\usepackage{newfloat}
\usepackage{listings}
\DeclareCaptionStyle{ruled}{labelfont=normalfont,labelsep=colon,strut=off} 
\floatstyle{ruled}
\newfloat{listing}{tb}{lst}{}
\floatname{listing}{Listing}
\title{Chatting about Upper-Body Expressive Human Pose and Shape Estimation}
\author{
    Yuxiang Zhao\textsuperscript{\rm 1,\rm 2}, Wei Huang\textsuperscript{\rm 1}\thanks{Corresponding author.}, Yujie Song\textsuperscript{\rm 1}, Liu Wang\textsuperscript{\rm 1}, Huan Zhao\textsuperscript{\rm 1}
}
\affiliations{
    \textsuperscript{\rm 1}Sun Yat-sen University
    \textsuperscript{\rm 2}Alibaba Group\\

    zhaoyao.zyx@alibaba-inc.com \\
    \texttt{https://zeusyux.github.io/CoEvoer}
}

\usepackage{bibentry}

\begin{document}
\maketitle
\begin{strip}
\vspace{-3em} 
\centering
  \includegraphics[width=1.0\linewidth]{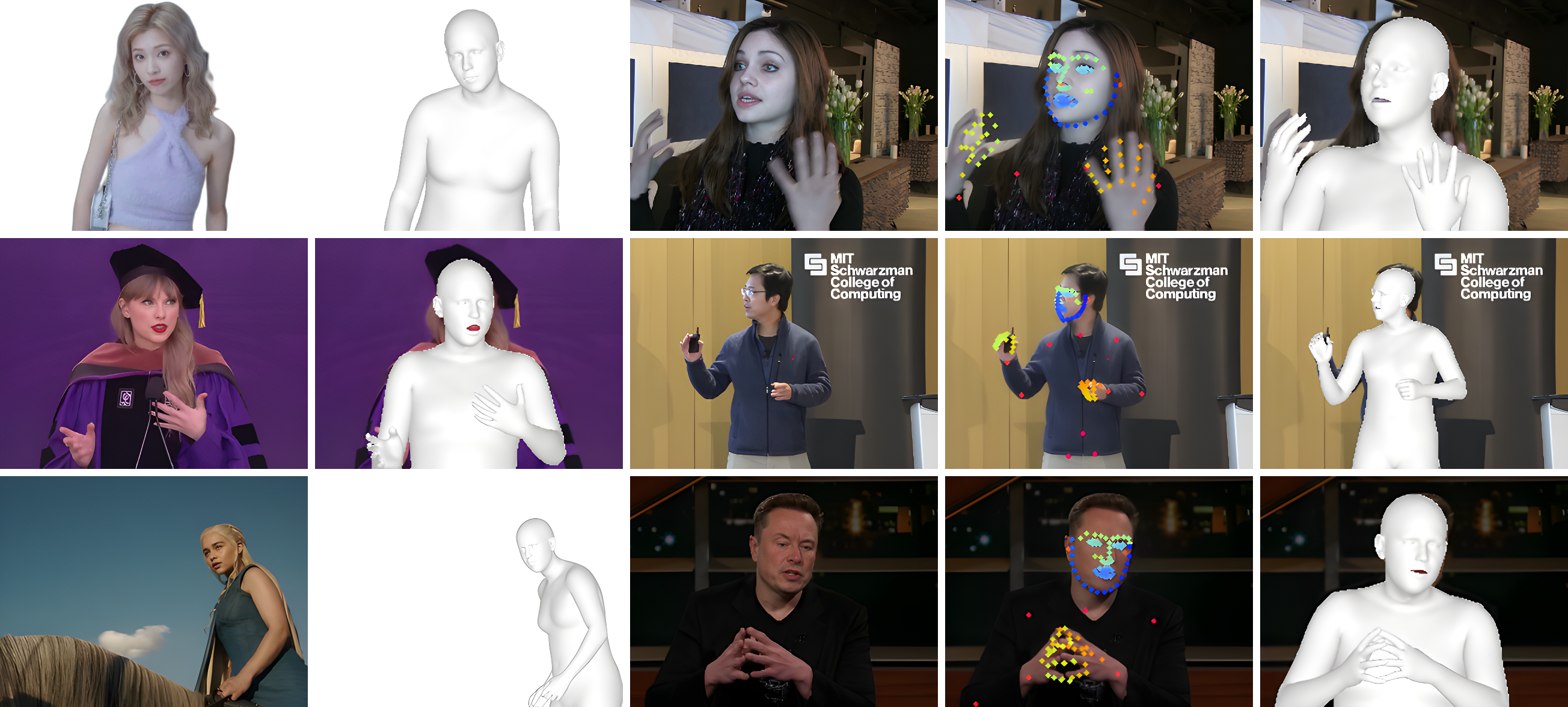}
\captionof{figure}{%
Our proposed CoEvoer achieves the mutual adaptation among different body parts through the mutual complementation and correction of global semantics and local features. When it is difficult to estimate some challenging regions like face and hands, it can rely on the semantic information provided by other parts for correction, which effectively improves the expressiveness and robustness of the method.
}
\label{Fig.1}
\end{strip}

\begin{abstract}
Expressive Human Pose and Shape Estimation (EHPS) plays a crucial role in various AR/VR applications and has witnessed significant progress in recent years. However, current state-of-the-art methods still struggle with accurate parameter estimation for facial and hand regions and exhibit limited generalization to wild images. To address these challenges, we present CoEvoer, a novel one-stage synergistic cross-dependency transformer framework tailored for upper-body EHPS. CoEvoer enables explicit feature-level interaction across different body parts, allowing for mutual enhancement through contextual information exchange. Specifically, larger and more easily estimated regions such as the torso provide global semantics and positional priors to guide the estimation of finer, more complex regions like the face and hands. Conversely, the localized details captured in facial and hand regions help refine and calibrate adjacent body parts. To the best of our knowledge, CoEvoer is the first framework designed specifically for upper-body EHPS, with the goal of capturing the strong coupling and semantic dependencies among the face, hands, and torso through joint parameter regression. Extensive experiments demonstrate that CoEvoer achieves state-of-the-art performance on upper-body benchmarks and exhibits strong generalization capability even on unseen wild images.
\end{abstract}

\section{Introduction}
Expressive human pose and shape estimation (EHPS) \cite{pang2023towards, liu2024deep, tian2023recovering, shen2024hmr} plays a central role in human behaviors understanding and has extensive applications \cite{zhang2024motiondiffuse, zhang2023remodiffuse, cai2022humman, hong2021garment4d, hong2022avatarclip} in virtual reality, motion capture and human-computer interaction. Given that many practical EHPS applications—such as online education and video conference—are inherently upper-body dominated, there has been increasing interest in developing methods specifically tailored to such scenarios, especially following the emergence of pioneering work \cite{lin2023one}.

Despite notable progress, accurately reconstructing expressive regions, particularly the face, hands, and their transition areas, remains a significant challenge \cite{choutas2020monocular, moon2022accurate, lin2023one, cai2023smpler}. Multi-stage methods \cite{moon2022accurate, choutas2020monocular} typically enhance detail recovery by cropping and upsampling specific regions, then feeding them into specialized expert models. While effective, these methods suffer from complex designs and prohibitive computational overhead, limiting their scalability in real-world applications. Frameworks such as OSX \cite{lin2023one} and SMPLer-X \cite{cai2023smpler} do not require separate expert networks for each part. These methods estimate the face, hands, and body jointly within a one-stage framework, offering improved inference efficiency. However, they typically treat different body parts independently in feature space, lacking explicit mechanisms for inter-part interaction. Consequently, they are unable to autonomously model the topological and semantic relationships among different human body components, leading to representational bottlenecks—particularly in the upper body, where fine-grained coordination is critical for expressive motion.

We argue that explicitly modeling feature-level interactions across body parts is crucial for expressive mesh recovery, particularly in upper-body-dominated scenes, for the following reasons: First, in upper-body scenarios, strong spatial and semantic dependencies exist among the body, hands, and face. These parts often move in a highly coordinated manner. For instance, at the spatial level, hand gestures are often conditioned on the configurations of proximal joints such as the shoulder and upper arm, while head and neck pose estimation is heavily influenced by facial orientation. At the semantic level, high-level human intent, such as making a phone call, typically induces characteristic postural patterns, including a downward head tilt and the placement of the hand near the ear. However, these models tokenize input into semantically agnostic patches, hindering their ability to capture structured priors—e.g., that hand motions are guided by arm configuration, or that facial direction should align with torso orientation. As a result, these models struggle to learn the topological and semantic relationships among human body components. Second, to simplify architecture and avoid training multiple expert models, current one-stage methods often crop features around the face and hands from shared feature maps. However, this inevitably introduces information loss—particularly with respect to the spatial context of hands. Incorporating inter-part interactions allows the model to recover lost spatial cues using global body posture, thereby enabling more accurate localization of hands and face. Moreover, such interactions improve robustness and generalization in challenging conditions such as occlusion or in-the-wild imagery: when either the body or hands/face is estimated with higher confidence, it can provide informative context to guide the estimation of the other. These observations motivate the need for a unified framework that can dynamically model semantic and spatial interactions across expressive body regions.

\begin{figure*}[tbp]
  \centering
   \includegraphics[width=1.0\linewidth]{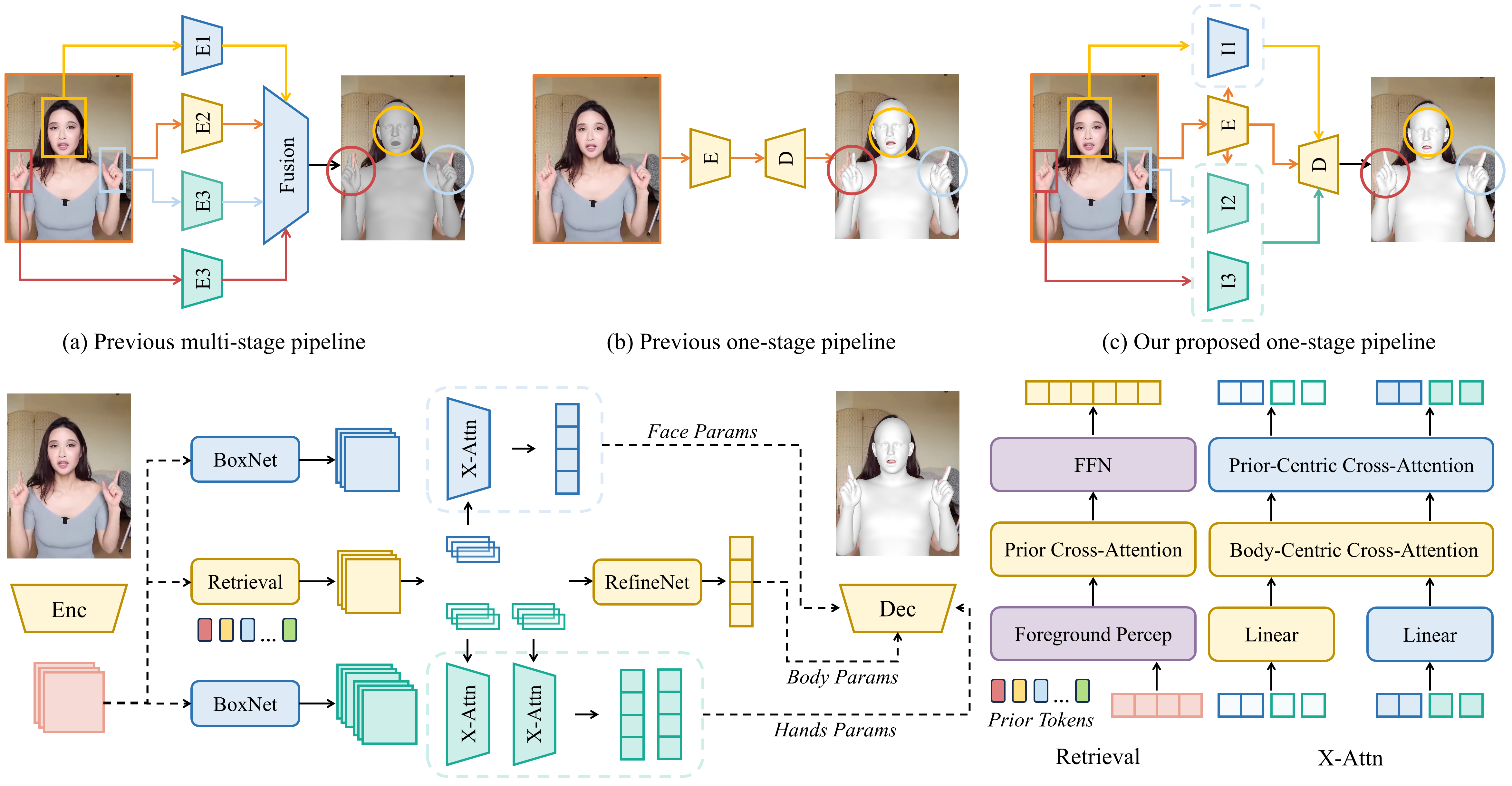}
   \caption{Comparison of existing mesh recovery methods and ours: multi-stage frameworks use part-specific experts (face E1, body E2, hands E3) for independent processing; one-stage frameworks adopt unified encoding-decoding, but still perform separate regressions for different parts without interaction. Our method preserves one-stage efficiency and adds explicit cross-part token-level interactions (I1: body-head, I2: body-left hand, I3: body-right hand).}
   \label{Fig.2}
\end{figure*}

In this work, we propose CoEvoer, a simple yet effective one-stage framework that facilitates token-level cross-part interaction, tailored for upper-body scenarios where the face, body, and hands are highly interdependent. Unlike existing one-stage transformer-based architectures that lack explicit modeling of inter-part semantics, our method incorporates structured cross-part communication to enhance representation learning. Specifically, facial expression tokens attend to upper-body pose embeddings to refine head orientation estimation, while hand-related tokens are contextually enriched via interactions with the shoulder and upper-arm features. Conversely, body representations are refined through feedback from facial and hand tokens, enabling a more holistic understanding of the full upper-body configuration. This explicit modeling of semantic dependencies across parts leads to more accurate human pose estimation, especially in challenging regions such as the face and hands, and helps correct anatomically implausible or incoherent hand poses. Moreover, the proposed interaction mechanism allows information from one part to compensate for occlusions or ambiguities in another. For instance, when the face is partially occluded, the head pose can still be reliably estimated using cues from the torso orientation or shoulder alignment. The comparison between our proposed framework and existing methods is illustrated in Figure \ref{Fig.2}. Our contributions can be summarized as follows.
\begin{itemize}
\item We propose a novel one-stage framework with explicit token-level interaction, tailored for upper-body scenarios where body parts exhibit strong semantic and spatial coupling. To the best of our knowledge, this is the first work to explicitly optimize cross-part interaction mechanisms for upper-body-focused EHPS.
\item By enabling semantic collaboration and contextual enhancement across body parts, our approach effectively captures inter-part dependencies and significantly improves model robustness and generalization, especially under occlusions and in-the-wild conditions.
\item Extensive experiments on public benchmarks demonstrate that our method achieves state-of-the-art performance on upper-body EHPS tasks, with notable improvements in pose estimation of challenging regions such as the face and hands.
\end{itemize}

\section{Related Work}
\label{sec:related}

\subsection{Expressive Human Pose and Shape Estimation}
EHPS not only needs to capture human postures and shapes \cite{li2022cliff, wang2023learning, wang2023zolly, kocabas2020vibe, kocabas2021pare, kolotouros2019learning}, but also to estimate facial expressions \cite{deng2019accurate, tewari2017mofa} and hand gestures \cite{sun2022learning, zhou2021monocular, xiang2019monocular}. In recent years, there has been a notable upsurge in research interests \cite{cai2023smpler, sun2024aios, baradel2024multi} focused on whole-body mesh recovery from monocular images. This trend is, to a certain extent, propelled by the ground-breaking research \cite{pavlakos2019expressive} in the domain of enhancing whole-body parametric models. Different from existing studies \cite{moon2022accurate, zhou2021monocular, feng2021collaborative} that separately recover the face, hands, or body, expressive human mesh recovery focuses on the joint estimation of the human face \cite{aldrian2012inverse, tewari2017mofa, deng2019accurate, egger20203d}, hands \cite{boukhayma20193d, chatzis2020comprehensive, huang2021survey}, and body \cite{choi2020pose2mesh, kanazawa2018end, kolotouros2019learning, kocabas2021pare, kocabas2021spec, zeng2022deciwatch, zeng2022smoothnet}. Since the face and hands regions are small but have a large number of details, many existing multi-stage methods crop out different body parts, scale them up to a higher resolution, and then feed them into their respective expert models for parameter estimation. ExPose \cite{choutas2020monocular} extracts high resolution cropped regions of the face and hands through the body driven attention mechanism and feeds them into their respective expert networks to utilize the specific knowledge from the face and hand datasets. FrankMocap \cite{rong2021frankmocap} operates by separately conducting 3D mesh recovery for the face, hands, and body. Afterward, the outputs are combined via an integration module. PIXIE \cite{feng2021collaborative} combines separate estimations by leveraging the shared shape space of SMPL-X that encompasses all body parts. Hand4Whole \cite{moon2022accurate} predicts the 3D wrists by exploiting MCP features, generating more accurate rotations and improving hand estimation. Such multi-stage frameworks are complex and have relatively long inference time, making it difficult to be applied in practical scenarios.

\subsection{One-stage Human Pose and Shape Estimation}
In recent years, with the continuous development of whole-body datasets, many one-stage frameworks \cite{lin2023one, cai2023smpler, baradel2024multi, sun2024aios} have been proposed for more concise and efficient expressive whole-body recovery. OSX \cite{lin2023one} is a pioneering work in one-stage frameworks. It proposes a simple yet effective transformer architecture, which significantly improves the model's inference efficiency. SMPLer-X \cite{cai2023smpler} leverages the large-scale model and further expands the dataset, endowing the model with stronger expressive power and transferability. Multi-HMR \cite{baradel2024multi} detects people by predicting 2D heatmaps of human positions, enabling the recovery of 3D human body meshes of multi-person from a single RGB image. Built upon DETR, AiOS \cite{sun2024aios} formulates the multi-person whole-body mesh recovery task as a progressive set prediction problem with diverse sequential detections. It is particularly suitable for scenarios that are crowded and feature significant occlusion. Despite their efficiency, most one-stage methods still treat each body part independently during encoding and decoding, lacking mechanisms for inter-part communication. This leads to spatial misalignment and limits the model’s ability to capture semantic correlations across regions—especially problematic in expressive upper-body scenarios where face, hands, and torso often move in a coordinated manner. Our work addresses these limitations by introducing a token-level interaction mechanism that enables contextual refinement across body parts within a one-stage framework.
\section{Preliminaries}
\label{sec:pre}
Our model is constructed based on the 3D parametric model SMPL-X \cite{pavlakos2019expressive}. Given the image $\mathcal{I} \in \mathbb{R} ^{H\times W\times 3}$, our proposed method estimates pose parameters $\theta \in \mathbb{R} ^{53\times 3}$ including body poses $\theta _{body} \in \mathbb{R} ^{22\times 3}$, left hand poses $\theta _{lhand} \in \mathbb{R} ^{15\times 3}$, right hand poses $\theta _{rhand} \in \mathbb{R} ^{15\times 3}$, jaw poses $\theta _{jaw} \in \mathbb{R} ^{1\times 3}$, as well as shape parameters $\beta \in \mathbb{R} ^{10}$ and facial expression parameters $\phi \in \mathbb{R} ^{10}$. Finally, these parameters are fed into a SMPL-X layer to obtain the final 3D human mesh.

\begin{figure*}[t]
  \centering
  \includegraphics[width=1.0\textwidth]{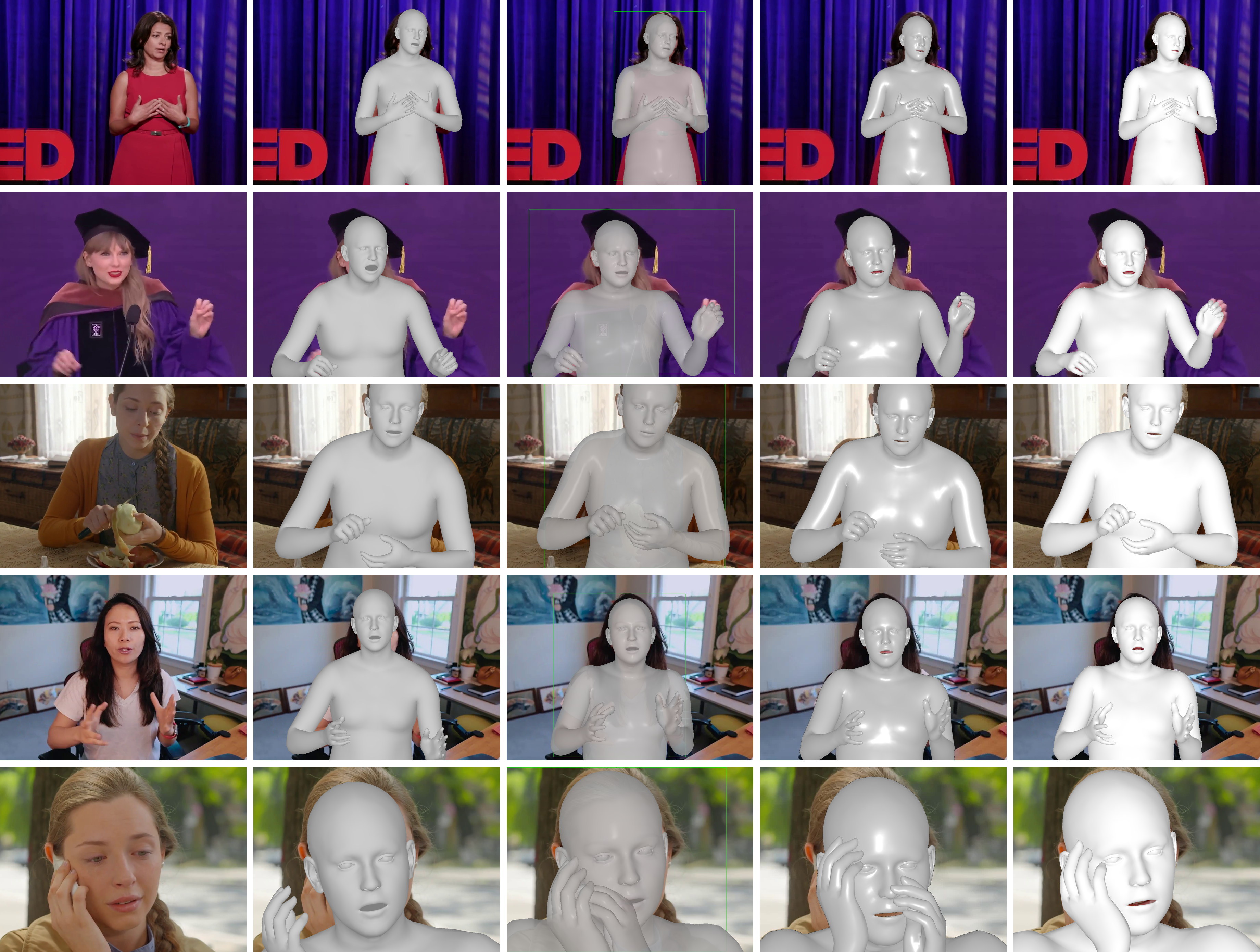}
  \caption{Qualitative visualization comparison on the UBody dataset. Each row shows, from left to right: the input image, results from Hand4Whole, AiOS, SMPLer-X, and our proposed CoEvoer. Best viewed in color with zoom-in for details.}
  \label{Fig.3}
\end{figure*}
\section{Proposed Method}
\subsection{Portrait Foreground Extraction}
In expressive upper-body pose estimation, performance often degrades when the subject appears distant from the camera. In such cases, detailed regions such as the face and hands occupy fewer pixels and are more vulnerable to background interference. This introduces irrelevant contextual information into body features, thereby hindering effective feature interaction across body parts. To mitigate this issue, we propose Portrait Foreground Extraction, a module designed to explicitly enhance the model’s focus on foreground regions. Inspired by recent advances in semantic segmentation \cite{hou2020strip, guo2022segnext, ni2024context}, Portrait Foreground Extraction isolates the human foreground by combining structured global context with refined local details. Given an input image $\mathcal{I}_{img}$, we first apply a convolutional encoder to extract initial features $\mathcal{Z}_{img}$ and adjust their channel dimensions. To capture the spatial distribution of the foreground, we perform adaptive pooling along the height and width dimensions, producing directional context features $\mathcal{Z}_{horizontal}$ and $\mathcal{Z}_{vertical}$, respectively. These are fused via element-wise addition to generate a coarse foreground activation map $\mathcal{Z}_{rec}$. To further refine this map, we apply strip convolutions in both horizontal and vertical directions. These operations enhance linear structures and boundary information, which are critical for distinguishing human silhouettes from background clutter. Finally, the refined foreground features are aggregated using a set of initialized queries, which act as foreground-aware tokens. Since these queries are subsequently used to inject pose priors into the interaction process for facial and hand pose estimation, we refer to them as Prior Interaction Tokens. The resulting foreground representation $\mathcal{Z}_{human}$ provides cleaner semantics and serves as a more robust foundation for downstream token-level feature interactions.

\subsection{Collaborative Evolution Enhancer}
\paragraph{Retrieval of different body parts.} To ensure inference efficiency, we adopt a streamlined one-stage architecture with a single shared encoder \cite{dosovitskiy2020image, li2022exploring, zhu2020deformable}. The feature representations for the face ($\mathcal{Z}_{face}$), left hand ($\mathcal{Z}_{lhand}$), and right hand ($\mathcal{Z}_{rhand}$) are extracted using BoxNet, which first predicts bounding boxes via multi-layer perceptron, followed by RoIAlign applied to the shared feature map to retrieve the corresponding local features. In upper-body scenarios, different body parts exhibit strong spatial and semantic correlations. Spatially, head movement is closely linked to the neck and shoulder regions, while hand positions are largely governed by arm configurations. Semantically, actions such as answering a phone call or gesturing during a conversation involve coordinated movements among the head, hands, and torso. These patterns suggest the existence of consistent inter-part dependencies. To model these relationships, we divide the torso features $\mathcal{Z}_{human}$—enhanced by the Portrait Foreground Extraction—into three tokens: $\mathcal{T}_{face}$, $\mathcal{T}_{lhand}$, and $\mathcal{T}_{rhand}$. These tokens serve to model the spatial and semantic dependencies between the torso and each corresponding body part, providing contextual cues that enhance the effectiveness of subsequent token-level feature interactions.

\paragraph{Explicit token-level cross-part interaction.} The suboptimal performance of one-stage frameworks in facial and hand pose estimation largely stems from their design for high inference efficiency. Unlike multi-stage approaches that crop high-resolution patches and process them with specialized branches, one-stage methods retrieve features directly from shared feature maps. As a result, the representations for the face, left hand, and right hand suffer from degraded spatial localization and become decoupled from the holistic body context. To address this limitation, we introduce a bidirectional cross-attention mechanism that explicitly models spatial and semantic dependencies among different body parts. The globally contextualized torso features—enhanced by the Portrait Foreground Extraction module—are partitioned into three segments: $\mathcal{T}_{face}$, $\mathcal{T}_{lhand}$, and $\mathcal{T}_{rhand}$, corresponding to the face, left hand, and right hand, respectively. These segments serve as keys and values to augment the queries from the facial and hand tokens. Conversely, the local part tokens also act as queries to retrieve complementary contextual cues from their respective torso segments, enabling mutual refinement between local and global representations. Specifically, the bidirectional interaction is defined as follows:
\begin{equation}
\widetilde{\mathcal{Z}}_{face} = \mathcal{F}^{corr}(\mathcal{Z}_{face}, \mathcal{T}_{face}, \mathcal{T}_{face})
\end{equation}
\begin{equation}
\widetilde{\mathcal{T}}_{face} = \mathcal{F}^{corr}(\mathcal{T}_{face}, \mathcal{Z}_{face}, \mathcal{Z}_{face})
\end{equation}
where $\widetilde{\mathcal{Z}}_{face}$ denotes the corrected facial feature, and $\widetilde{\mathcal{T}}_{face}$ is the refined torso segment associated with the face. The general form of the attention-based correction function is:
\begin{equation}
\mathcal{F}^{corr}(Q, K, V) = \mathcal{F}^{fc}(\mathcal{F}^{attn}(Q, K, V))
\end{equation}
The same interaction procedure is applied to the left hand and right hand tokens and their corresponding torso segments, namely $(\mathcal{Z}_{lhand}, \mathcal{T}_{lhand})$ and $(\mathcal{Z}_{rhand}, \mathcal{T}_{rhand})$. To consolidate the enhanced global context, we further fuse the refined torso segments $\widetilde{\mathcal{T}}_{face}$, $\widetilde{\mathcal{T}}_{lhand}$, and $\widetilde{\mathcal{T}}_{rhand}$ using a multi-layer perceptron, referred to as the RefineNet module. This fusion propagates part-aware semantics into the global body representation, reinforcing anatomical coherence. By leveraging bidirectional cross-attention for token-level feature refinement, our method yields more accurate and anatomically plausible full-body pose estimations. This mechanism also enhances the localization of facial and hand keypoints and effectively mitigates issues such as unnatural hand configurations. Finally, we feed the refined part-specific features into separate MLP heads to predict the output parameters: $\mathcal{O}_{body}$, $\mathcal{O}_{face}$, $\mathcal{O}_{lhand}$, and $\mathcal{O}_{rhand}$.

\paragraph{Loss function.} Our proposed CoEvoer is trained in an end-to-end manner by minimizing the loss function $\mathcal{L}$, defined as follows:
\begin{equation}
\mathcal{L}=\mathcal{L} _{param}+\mathcal{L} _{kpts}+\mathcal{L} _{bbox}
\end{equation}
where $\mathcal{L}_{param}$ is the L1 distance between the estimated SMPL-X parameters and the ground truth, and $\mathcal{L}_{kpts}$ is the L1 distance between the predicted keypoints and the ground truth, which includes the 3D keypoints of the human body and the projected 2D keypoints. $\mathcal{L}_{param}$ and $\mathcal{L}_{kpts}$ are used to supervise the human body posture and shape. $\mathcal{L}_{bbox}$ refers to the estimated coordinates of the facial and hand bounding boxes, which are used to supervise the detection of the facial and hand regions.
\section{Experiments}
\subsection{Experimental Setup}
Following existing methods, we use Human3.6M \cite{ionescu2013human3}, COCO-Wholebody \cite{lin2014microsoft}, and MPII \cite{andriluka20142d} as the training set. The SMPL/SMPL-X pseudo-GTs are obtained from EFT \cite{joo2021exemplar} and NeuralAnnot \cite{moon2022neuralannot}. Since UBody \cite{lin2023one} is the most recent large-scale dataset for human pose and shape estimation, offering significantly more data and a broader coverage of real-world scenarios than previous datasets, it has been widely adopted as a representative benchmark for evaluating the downstream applicability of human mesh recovery methods. In our experiments, we fine-tune our model on the UBody and evaluate its performance accordingly.
\paragraph{Implementation.} PyTorch is used for implementation. We use the Adam optimizer to conduct the training with an initial learning rate of $1\times 10^{-4}$. Meanwhile, we adopt data augmentation methods including scaling, rotation, random horizontal flip, and color jittering. All the hyperparameter settings are kept consistent with comparative methods.
\subsection{Qualitative Comparison with SOTA}
As depicted in Figure \ref{Fig.3}, we present a qualitative comparison between our proposed CoEvoer, and existing state-of-the-art approaches, including the two-stage pipeline Hand4Whole and the one-stage methods SMPLer-X and AiOS. Notably, both SMPLer-X and AiOS are trained with additional external datasets. In the first example, both Hand4Whole and SMPLer-X suffer from noticeable finger interpenetration artifacts, while AiOS fails to accurately estimate the facial and hand poses. In the second example, Hand4Whole and SMPLer-X exhibit suboptimal hand recovery, and AiOS produces anatomically implausible hand configurations. For the remaining examples, we omit detailed discussion and encourage readers to evaluate the results based on global body consistency, self-intersections, and unnatural pose artifacts. It is particularly worth noting that in the fifth and most challenging case, all baseline methods exhibit substantial estimation failures, whereas CoEvoer maintains robust and accurate performance.
\paragraph{Evaluation metrics.} For expressive human pose and shape estimation, we adopt the mean per-vertex position error (MPVPE) as the main evaluation metric. Meanwhile, we also report the procrustes analysis mean per-vertex position error (PA-MPVPE) after rigid alignment to further analyze the performance of the model. Regarding the 3D hand error, we report the average 3D errors of the left hand and the right hand. The units of all the reported metrics are millimeters.
\subsection{Quantitative Comparison with SOTA}
Table \ref{table1} shows the detailed comparison between our proposed CoEvoer and existing human pose and shape estimation methods. We achieved improvements of 10.1\% and 16.9\% in MPVPE and PA-MPVPE respectively on UBody, among which the MPVPE of the hand achieved an improvement of 28.7\%. It is worth noting that even though SMPLer-X and AiOS used additional datasets, CoEvoer still outperforms the existing SOTA methods. As a unified one-stage framework, our method outperforms even the two-stage pipelines without relying on any supplementary face-only or hand-only datasets. As shown in Table \ref{table2} and Table \ref{table3}, we also achieved performance improvements of 4.6\% and 6.4\% on AGORA and EHF respectively, and the estimation results of the hand and face on the EHF dataset were improved by 8.9\% and 5.3\% respectively.

\begin{table}[htbp]
    \centering
    \scalebox{0.89}{
    \begin{tabular}{l|ccc|ccc}
        \toprule
        Method & \multicolumn{3}{c|}{MPVPE $\downarrow$} & \multicolumn{3}{c}{PA-MPVPE $\downarrow$} \\
        \cmidrule(lr){2-4} \cmidrule(lr){5-7} 
         & All & Hand & Face & All & Hand & Face \\
        \midrule
        ExPose & 171.5 & 83.7 & 45.1 & 66.9 & 12.0 & 3.9 \\
        PIXIE & 168.4 & 55.6 & 45.2 & 61.7 & 12.2 & 4.2 \\
        H4W & 104.1 & 45.7 & 27.0 & 44.8 & 8.9 & 2.8 \\
        OSX & 81.9 & 41.5 & 21.2 & 42.2 & 8.6 & \cellcolor{lightyellow}2.0 \\
        AiOS & 58.6 & \cellcolor{lightyellow}39.0 & \cellcolor{lightyellow}19.6 & 32.5 & \cellcolor{lightyellow}7.3 & 2.8 \\
        SMPLer-X & \cellcolor{lightyellow}57.4 & 40.2 & 21.6 & \cellcolor{lightyellow}31.9 & 10.3 & 2.8 \\
        \midrule
        Ours & \cellcolor{lightpink}51.6 & \cellcolor{lightpink}27.8 & \cellcolor{lightpink}16.2 & \cellcolor{lightpink}26.5 & \cellcolor{lightpink}7.1 & \cellcolor{lightpink}1.8 \\
        \bottomrule
    \end{tabular}
    }
    \caption{Reconstruction errors on UBody. \colorbox{lightpink}{Red} background indicates best results, \colorbox{lightyellow}{yellow} background indicates second best results.}
    \label{table1}
\end{table}

\begin{table}[htbp]
    \centering
    \scalebox{0.89}{
    \begin{tabular}{l|ccc|ccc}
        \toprule
        Method & \multicolumn{3}{c|}{MPVPE $\downarrow$} & \multicolumn{3}{c}{PA-MPVPE $\downarrow$} \\
        \cmidrule(lr){2-4} \cmidrule(lr){5-7} 
         & All & Hand & Face & All & Hand & Face \\
        \midrule
        ExPose & 219.8 & 115.4 & 103.5 & 88.0 & 12.1 & 4.8 \\
        FrankMocap & 218.0 & 95.2 & 105.4 & 90.6 & 11.2 & 4.9 \\
        PIXIE & 203.0 & 89.9 & 95.4 & 82.7 & 12.8 & 5.4 \\
        H4W & 183.9 & 72.8 & 81.6 & 73.2 & \cellcolor{lightpink}9.7 & \cellcolor{lightpink}4.7 \\
        OSX & \cellcolor{lightyellow}168.6 & \cellcolor{lightyellow}70.6 & \cellcolor{lightyellow}77.2 & \cellcolor{lightyellow}69.4 & 11.5 & 4.8 \\
        \midrule
        Ours & \cellcolor{lightpink}160.8 & \cellcolor{lightpink}68.7 & \cellcolor{lightpink}75.7 & \cellcolor{lightpink}65.9 & \cellcolor{lightyellow}10.1 & \cellcolor{lightpink}4.7 \\
        \bottomrule
    \end{tabular}
    }
    \caption{Reconstruction errors on AGORA. \colorbox{lightpink}{Red} background indicates best results, \colorbox{lightyellow}{yellow} background indicates second best results.}
    \label{table2}
\end{table}
\begin{table}[htbp]
    \centering
    \scalebox{0.89}{
    \begin{tabular}{l|ccc|ccc}
        \toprule
        Method & \multicolumn{3}{c|}{MPVPE $\downarrow$} & \multicolumn{3}{c}{PA-MPVPE $\downarrow$} \\
        \cmidrule(lr){2-4} \cmidrule(lr){5-7} 
         & All & Hand & Face & All & Hand & Face \\
        \midrule
        ExPose & 77.1 & 51.6 & 35.0 & 54.5 & 12.8 & \cellcolor{lightyellow}5.8 \\
        FrankMocap & 107.6 & \cellcolor{lightyellow}42.8 & - & 57.5 & 12.6 & - \\
        PIXIE & 89.2 & \cellcolor{lightyellow}42.8 & 32.7 & 55.0 & \cellcolor{lightyellow}11.1 & \cellcolor{lightpink}4.6 \\
        H4W & 76.8 & \cellcolor{lightpink}39.8 & \cellcolor{lightyellow}26.1 & 50.3 & \cellcolor{lightpink}10.8 & \cellcolor{lightyellow}5.8 \\
        OSX & \cellcolor{lightyellow}70.8 & 53.7 & 26.4 & \cellcolor{lightyellow}48.7 & 15.9 & 6.0 \\
        \midrule
        Ours & \cellcolor{lightpink}66.3 & 48.9 & \cellcolor{lightpink}25.0 & \cellcolor{lightpink}46.4 & 14.8 & \cellcolor{lightyellow}5.8 \\
        \bottomrule
    \end{tabular}
    }
    \caption{Reconstruction errors on EHF. \colorbox{lightpink}{Red} background indicates best results, \colorbox{lightyellow}{yellow} background indicates second best results.}
    \label{table3}
\end{table}
\subsection{Ablation Study and Discussion}
As shown in Table \ref{table4}, we design two variants to evaluate the effectiveness of our proposed modules. Variant1 is designed to assess the impact of explicitly modeling token-level interactions across different body parts. It performs feature fusion via self-attention on concatenated part features without using cross-attention for inter-part information exchange, thereby isolating the contribution of explicit cross-part communication in our framework.

\begin{figure}[htbp]
  \centering
   \includegraphics[width=0.15\textwidth]{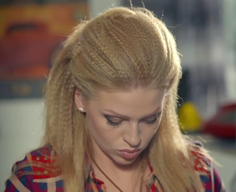}
   \includegraphics[width=0.15\textwidth]{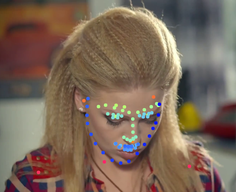}
   \includegraphics[width=0.15\textwidth]{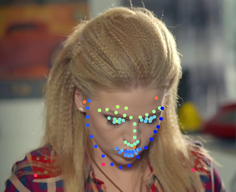}
   \caption{Comparison of facial keypoint estimation results. The first column shows the input image, the second column visualizes the output of Variant1, and the third column shows the result produced by our proposed CoEvoer. Best viewed in color and zoom-in for more clarity.}
   \label{Fig.4}
\end{figure}

To further demonstrate the effectiveness of our approach, we present a visualization of keypoint estimation results for CoEvoer and the variant in Figure \ref{Fig.4}, focusing on facial and partial torso regions, including the neck and shoulders. As illustrated, the subject in the image is looking downward, causing the facial keypoints to appear significantly lower than in typical upright poses. The variant tends to predict relatively accurate torso keypoints, which are less affected by the head orientation, while its facial estimation results are biased upwards. In contrast, our method captures the dependency between the torso and face through token-level interaction. By recognizing the subtle deviation in the neck position, CoEvoer successfully infers the downward head pose and accordingly corrects the facial keypoints, leading to a more accurate and coherent estimation.

\begin{figure}[htbp]
  \centering
   \includegraphics[width=0.15\textwidth]{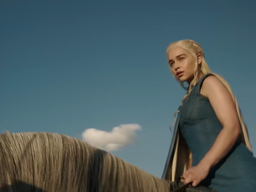}
   \includegraphics[width=0.15\textwidth]{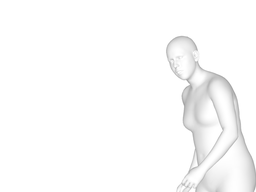}
   \includegraphics[width=0.15\textwidth]{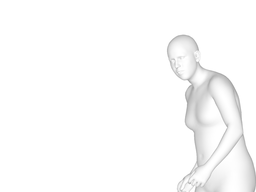}
   \caption{Comparison of mesh recovery results. The first column shows the input image, the second column visualizes the output of Variant1, and the third column shows the result produced by our proposed CoEvoer. Best viewed in color and zoom-in for more clarity.}
   \label{Fig.5}
\end{figure}

We further compare our proposed method with the variant under extreme conditions. As shown in Figure \ref{Fig.5}, the input is a wild image where the subject’s right arm is completely occluded. Our method accurately infers the approximate location of the invisible right arm and even captures the specific hand gesture of gripping the saddle while riding. In contrast, the variant model predicts a generic downward-facing hand pose, failing to reflect the contextual semantics of the scene.

\begin{table}[htbp]
    \centering
    \scalebox{0.85}{
    \begin{tabular}{l|ccc|ccc}
        \toprule
        Method & \multicolumn{3}{c|}{MPVPE $\downarrow$} & \multicolumn{3}{c}{PA-MPVPE $\downarrow$} \\
        \cmidrule(lr){2-4} \cmidrule(lr){5-7} 
         & All & Hand & Face & All & Hand & Face \\
        \midrule
        Variant1-w/o C.E.E. & 66.5 & 36.9 & 23.8 & 33.0 & 8.9 & 2.1 \\
        Variant2-w/o P.F.E. & 54.4 & 29.7 & 17.7 & 27.5 & 9.4 & 2.1 \\
        Ours & 51.6 & 27.8 & 16.2 & 26.5 & 7.1 & 1.8 \\
        \bottomrule
    \end{tabular}
    }
    \caption{The ablation study on UBody. C.E.E. and P.F.E. are abbreviations for Collaborative Evolution Enhancer and Portrait Foreground Extraction respectively.}
    \label{table4}
\end{table}

\begin{figure}[htbp]
  \centering
   \includegraphics[width=1.0 \linewidth]{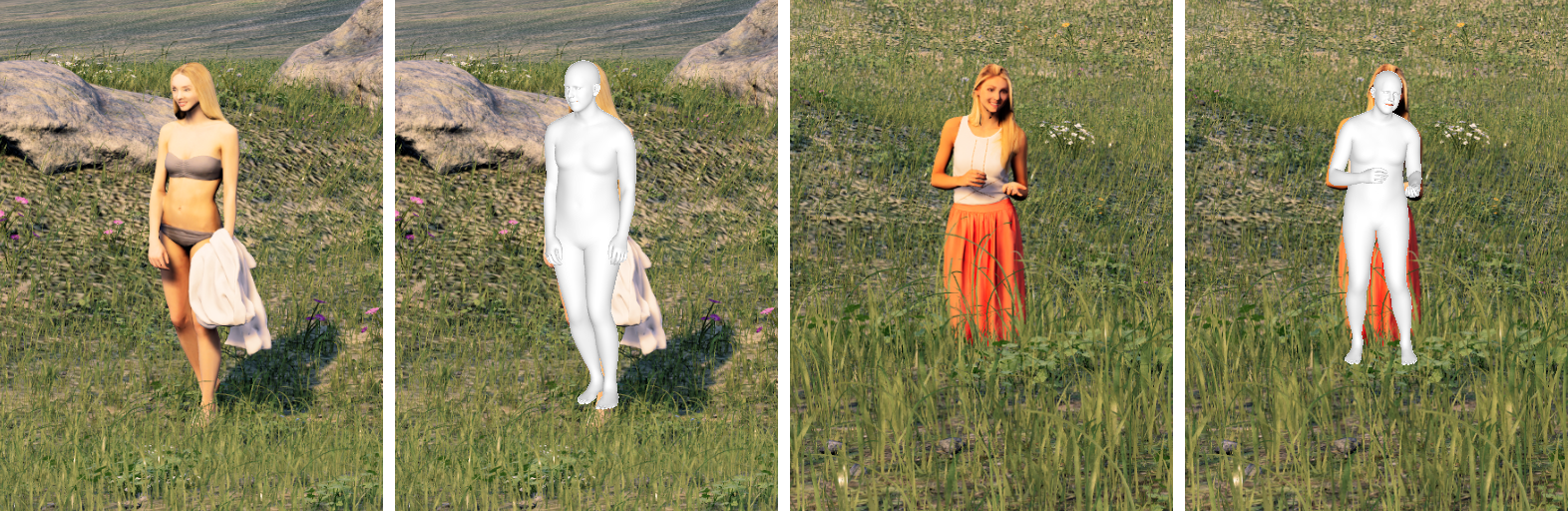}
   \caption{Qualitative visualization of human mesh recovery results on the AGORA dataset. The first and third columns show the input image, the second and fourth columns show the human mesh recovery produced by our proposed CoEvoer. Best viewed in color and zoom-in for more clarity.}
   \label{Fig.6}
\end{figure}

While the proposed CoEvoer framework is primarily tailored to improve expressive upper-body pose and shape estimation, the quantitative results in Table \ref{table2} and Table \ref{table3} indicate consistent improvements across full-body benchmarks as well. The qualitative visualizations in Figure \ref{Fig.6} further highlight the robustness of our approach under challenging scenarios, such as severe hand occlusions (left) and uncommon pose configurations (right). It is worth noting that, in the left example, the reconstructed lower limbs exhibit reduced accuracy. We attribute this phenomenon to the relatively weak correlation between the legs and the face/hands in full-body settings, which limits the amount of useful information that can be inferred for lower-limb estimation from upper-body cues. We believe this observation, together with our experimental findings, can motivate future studies on inter-region dependencies in human pose and shape estimation.

\section{Conclusion}
In this work, we introduced CoEvoer, a simple yet effective one-stage framework for expressive human pose and shape estimation in upper-body-dominated scenarios. Unlike existing approaches that treat body parts independently, CoEvoer explicitly models token-level interactions across the face, hands, and body, capturing rich semantic and spatial dependencies inherent to expressive motion. Through structured cross-part communication, CoEvoer enables each region to leverage contextual cues from others, resulting in more coherent and anatomically plausible mesh recovery, especially in challenging regions such as the hands and face. Our method preserves the efficiency of one-stage architectures while significantly enhancing representational capacity and generalization, particularly under occlusion and in-the-wild conditions. Extensive experiments on public benchmarks demonstrate that CoEvoer achieves state-of-the-art performance, validating the efficacy of explicit inter-part interaction in upper-body EHPS tasks.

\section{Acknowledgments}
The work was supported by the Shenzhen Science and Technology Program (Grant No. JCYJ20230807110807015), the Guangdong Basic and Applied Basic Research Foundation (Grant No. 2025A1515011757), and Baidu, Inc. We thank the anonymous reviewers for their constructive comments and suggestions.

\bibliography{Formatting-Instructions-LaTeX-2026}
\end{document}